\documentclass{article}
\usepackage{spconf,amsmath,graphicx}

\usepackage{amsfonts,amssymb}
\usepackage{multirow}
\usepackage{adjustbox}
\usepackage{threeparttable}

\title{MULTI-LEVEL ADAPTIVE REGION OF INTEREST AND GRAPH LEARNING FOR\\FACIAL ACTION UNIT RECOGNITION}
\name{Jingwei Yan$^{1,2\dagger}$, Boyuan Jiang$^{2\dagger}$, Jingjing Wang$^{1}$, Qiang Li$^{1}$, Chunmao Wang$^{1}$, Shiliang Pu$^{1\star}$}
\address{$^{1}$ Hikvision Research Institute, China \qquad $^{2}$ Zhejiang University, China}
\begin{document}
%\ninept
%
\maketitle
\let\thefootnote\relax\footnotetext{$^{\dagger}$ Equal contribution.}
\let\thefootnote\relax\footnotetext{$^{\star}$ Corresponding author: Shiliang Pu (pushiliang.hri@hikvision.com).}
\let\thefootnote\relax\footnotetext{This work was supported by National Key Research and Development Project of China (2018YFC0807702).}
\begin{abstract}
In facial action unit (AU) recognition tasks, regional feature learning and AU relation modeling are two effective aspects which are worth exploring. However, the limited representation capacity of regional features makes it difficult for relation models to embed AU relationship knowledge. In this paper, we propose a novel multi-level adaptive ROI and graph learning (MARGL) framework to tackle this problem. Specifically, an adaptive ROI learning module is designed to automatically adjust the location and size of the predefined AU regions. Meanwhile, besides relationship between AUs, there exists strong relevance between regional features across multiple levels of the backbone network as level-wise features focus on different aspects of representation. In order to incorporate the intra-level AU relation and inter-level AU regional relevance simultaneously, a multi-level AU relation graph is constructed and graph convolution is performed to further enhance AU regional features of each level. Experiments on BP4D and DISFA demonstrate the proposed MARGL significantly outperforms the previous state-of-the-art methods.
\end{abstract}
\section{Introduction}
%Facial action units (AUs), which are defined to describe facial dynamic behaviors caused by the movement of one or a group of facial muscles~\cite{ekman1997face}, have been widely applied in human facial expression related tasks such as complex expression analysis, deception detection and animation production. As a fundamental part of these applications, AU recognition has drawn increasing attentions in affective computing and signal processing in the past decade~\cite{zhao2016deep,shao2018deep,niu2019multi}.

\begin{figure}[tbp]
	\begin{center}
		\includegraphics[width=0.7\columnwidth]{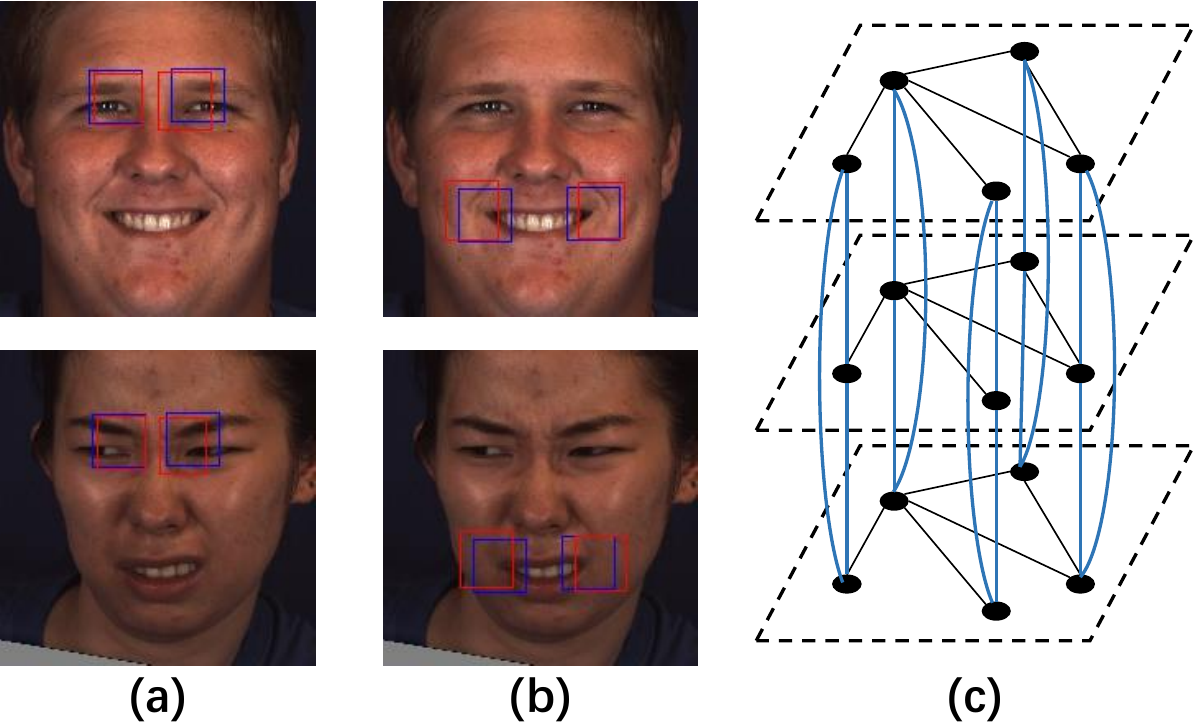}
	\end{center}
	\caption{(a) and (b) illustrate the adaptive ROIs of AU4 and AU12. Blue and red rectangles denote the initial and the adaptively adjusted ROIs. (c) The multi-level AU relation graph designed to enhance AU regional features by both intra-level AU relations and inter-level AU ROI relevance. Black and blue solid lines denote the intra and inter connections.}
	\label{fig1}
\end{figure}
% Bold dots denote the AU representations at multiple levels.

Each AU corresponds with one or a specific group of facial muscles~\cite{velusamy2011method}, which locates in a particular local region and contains unique structure or texture information. It is beneficial to consider these AU related local regions separately rather than treating them equally in order to obtain discriminative feature representations. Many existing works~\cite{li2017eac,shao2018deep,corneanu2018deep,zhang2019context} first cropped regions of interest (ROI) from facial image or intermediate features, and then employed independent convolutional neural networks (CNNs) for each ROI. However, the bounding box of each ROI is either predefined at fixed locations of facial image or determined based on the rough position relation between AU and facial landmarks. Meanwhile, the size of each ROI is identical. Thus the location and scale of the handcrafted ROI remain the same during training and inference processes, which ignores the fact that the scale of each AU varies from each other. As shown in Fig.~\ref{fig1} (a) and (b), the ROIs of AU4 (brow lowerer) and AU12 (lip corner puller) in the blue bounding boxes are of the same size and locate in predefined locations. Nevertheless, the inner part of brows beyond the blue boxes and the wrinkle caused by the lip movement also provide useful cues to recognize AU4 and AU12. It is helpful for these regions to be included in ROIs. %On the other hand, due to various face shapes and texture features, it is not accurate enough to determine ROI locations on basis of the same position relationship rule between AUs and landmarks for different individuals. %which could lead to mismatching problems occasionally.
%Similarly, the wrinkle caused by the lip movement is also helpful to recognize AU12.

Besides exploiting discriminative regional feature presentation, AU relationship modeling is often introduced to further boost the recognition performance~\cite{eleftheriadis2015multi,walecki2017deep,shao2019facial}. Some works utilized restricted Boltzmann machine (RBM)~\cite{wang2013capturing} to model the dependencies between AUs in a post-process manner, which was not end-to-end trainable. Recently, gated graph neural network (GGNN)~\cite{li2019} or graph convolutional network (GCN)~\cite{niu2019multi,liu2020relation} was tactfully incorporated so that prior AU relation knowledge can be leveraged to derive reliable predictions and the whole framework is end-to-end trainable. In these graphic models, each node represents regional features of a certain AU which is critical for subsequent graph inference. As low-level features of backbone network is AU semantically weak and high-level features lack subtle texture details, a straightforward solution is to add or concatenate them together~\cite{li2019}. However, the integrated features are still limited in representation capacity, which makes it difficult for graphic models to embed the relation knowledge effectively into the final features for prediction.
%Some groups of AUs are very probably to emerge simultaneously like AU6 (cheek raiser) and AU12 in a smile face, while on the contrary, some groups of AUs barely show up together due to facial anatomy restrictions, etc. It has been proven that incorporating such knowledge is beneficial to AU recognition.

Given the concerns above, we propose a new approach called multi-level adaptive ROI and graph learning (MARGL) to adaptively crop ROIs at multiple levels and simultaneously incorporate the intra-level AU relation and inter-level AU region relevance to enhance the discrimination of ROI features. In this framework, an adaptive ROI learning module is proposed to obtain precise ROIs which can match with different AUs and individuals automatically as illustrated in Fig.~\ref{fig1} (a) and (b). The module is plugged in different levels of the backbone network. Meanwhile, as shown in Fig.~\ref{fig1} (c), a unified multi-level AU relation graph is designed to integrate prior AU relation knowledge which is obtained based on the conditional probability between AUs calculated in the dataset, and ROI feature relations which connect the same AU across levels to facilitate the cross-level information fusion. Based on the multi-level graph, both intra-level AU relation and inter-level region relevance can be encoded in ROI features to further enhance the discrimination.

%The contribution of this paper is three-folded. I) We propose an adaptive ROI learning module which can easily be deployed in different levels of the backbone network and automatically adjust the position and size of each ROI to optimally crop the regional features. II) We propose a multi-level AU relation learning method to embed the intra-level AU relation and inter-level region relevance into regional features to further enhance the discrimination. III) Extensive experiments on two widely used benchmarks, i.e., BP4D and DISFA, demonstrate the superiority of the proposed MARGL. %In terms of F1-frame, improvements of 3.2\% and 7.8\% are achieved respectively.

\begin{figure*}[htbp]
	\begin{center}
		\includegraphics[width=0.7\textwidth]{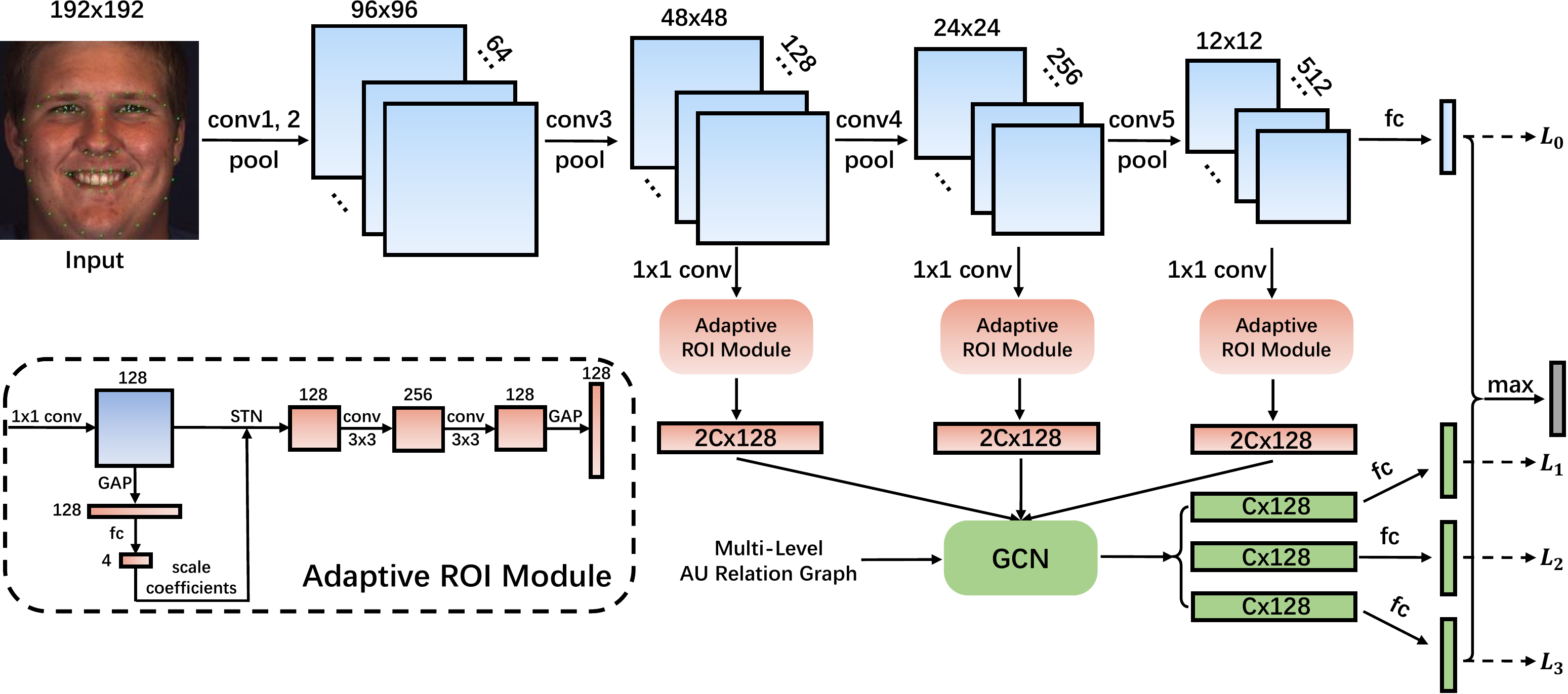}%width=5.7in,height=2.3in
	\end{center}
	\caption{The architecture of MARGL framework. Individual numbers indicate the channels of feature maps or feature dimension.}
	\label{framework}
\end{figure*}

\section{Proposed Method}
The proposed MARGL framework is illustrated in Fig.~\ref{framework}. It consists of three main modules, i.e., multi-level feature learning (backbone network), adaptive ROI learning, and multi-level AU relation embedding. %The multi-level features, from low-level texture details to high-level AU semantic information, are learned by the backbone CNN model. Individual adaptive ROI learning modules are employed in the feature maps of the last three levels to obtain discriminative ROI feature representation. Via graph convolution on the constructed multi-level AU relation graph, ROI features of multiple levels are further enhanced by encoding the intra-level AU relation and inter-level region relevance simultaneously.

\subsection{Adaptive ROI Learning}
The adaptive ROI learning module aims to obtain precise ROI locations of each AU for different individuals adaptively so as to capture discriminative regional feature representation. Different from previous works which kept the cropped region size fixed for all AUs during training and inference, the proposed adaptive ROI learning module can adjust the position and size of each ROI optimally in a data-driven manner.

Considering the symmetry of face regions, we divide the face into left and right parts. The initial $C$ AU centers on each side are obtained by the rough relation between AUs and landmarks~\cite{li2017eac}, where $C$ is the number of AUs. Then for each AU center, we map its coordinate from the original image to corresponding feature maps proportionally. The top-left and bottom-right coordinates of initial ROI whose size is $k\times k$ can thus be determined. In order to refine the ROI adaptively, we introduce a group of scale factors to adjust the two coordinates. The updated coordinate value is given as follows.
\begin{equation}
    \hat{p_i^j} = \beta_i^j*p_i^j,
\end{equation}
where $\beta_i^j$ is the scale factor to obtain the refined position $\hat{p_i^j}$. $(p_1^j,p_2^j)$ and $(p_3^j,p_4^j)$ are the original top-left and bottom-right coordinates respectively. $j\in [1,C]$ indicates the index of the AU. Therefore, with the scale factors, the bounding box of ROI can be located in any position if needed. In order to adaptively learn the scale factors from input data, we utilize a set of convolutional and fully connected layers to derive $\beta_i^j$ from the global feature maps where ROI features are cropped.

The structure of adaptive ROI learning module is shown in the dashed box in Fig.~\ref{framework}. In order to unify the shape of the input data, $1\times 1$ convolution is first applied to squeeze the channels of feature maps from different dimensions to 128. Based on the squeezed feature maps, global average pooling (GAP) is employed to reduce the computation cost. Then the scale factors are outputted with the fully connected layer. As the ROI coordinates after adjustment are fractional, the spatial transformer network (STN)~\cite{jaderberg2015spatial} is utilized to obtain the cropped features. The transformation matrix $\Theta$ of STN is formulated as follows.
\begin{equation}
    \Theta=\left[
    \begin{matrix}
      s_x & 0 & t_x \\
      0 & s_y & t_y
      \end{matrix}
    \right],
\end{equation}
where $s_x$ and $s_y$ are scaling coefficients, $t_x$ and $t_y$ are translation coefficients which can be calculated via the corresponding top-left and bottom-right coordinates easily. Once we obtain the target regional feature maps, independent regional learning networks which consist of two convolution layers and one GAP layer are applied to further refine the regional feature maps for each AU. Consequently, we obtain $2C$ adaptive ROI features totally due to the facial symmetry.
%$3\times 3$ Output channels of the two convolution layers are 256 and 128 respectively.

\subsection{Multi-Level AU Relation Embedding}
Correlations between AUs are often taken into consideration. Besides that, as level-wise features focus on different aspects of feature representation, there exists strong relevance between regional features across multiple levels of the backbone network. To leverage the intra-level relationship between AUs and inter-level AU regional relevance simultaneously, a multi-level AU relation graph is constructed to embed such relation knowledge into the ROI features.

%\subsubsection{Graph Definition}
The multi-level AU relation graph $G$ is shown in Fig.~\ref{fig1} (c). For intra-level, the sub-graph $G_0$ in dashed box describes the correlation between AUs which is shared by all three levels. For inter-level, only nodes of the same AU are connected to facilitate the information fusion across levels. The graph contain nodes from three levels, i.e., there are totally $3C$ nodes in the graph. To describe the graph $G$ with adjacency matrix $A$, we first calculate the adjacency matrix $A_0$ of the sub-graph $G_0$. The edge $e_{i,j}$ between AU $i$ and AU $j$ is obtained based on the conditional probability statistics in the dataset, thus $A_0(i,j)$ is defined as
%We define the multi-level AU relation graph as $G = (V,E)$, where $V$ is the node set and $E$ is the edge set. Each node $v_i^k\in V$ represents the ROI feature of the $i^{th}$ AU in the $k^{th}$ level and the edge indicates the pair-wise relationship between the two nodes.
\begin{gather}
\label{e}
A_0(i,j)=
\begin{cases}
1,& p(y_i=1|y_j=1)\ge p_{pos}\\
0,& else,
\end{cases}
\end{gather}
where $p_{pos}$ is a threshold to determine whether two AUs are connected in the graph, which is set to be 0.3. Then the adjacency matrix of graph $G$ is defined as follows.
\begin{equation}
    A=\left[
    \begin{matrix}
      A_0 & I & I \\
      I & A_0 & I \\
      I & I & A_0\end{matrix}
    \right],
\end{equation}
where $A_0$ is the same for all three levels which describes the relations between AUs. $I$ is the identity matrix of dimension $C$ which indicates the cross-level connection of the same AU in $G$. After the multi-level AU relation graph $G$ is built, we can perform reasoning on the graph with graph convolutional networks (GCN)~\cite{kipf2017semi} to encode the relationship knowledge into the ROI features. The process of the GCN can be viewed as a message-passing operation within the graph. Formally, one layer of GCN can be written as:

%Inspired by~\cite{velivckovic2017graph}, the connection weight between two related nodes in $G$ is calculated according to the semantic similarity between them as follows,
%\begin{equation}
%Sim(x_i,x_j)=\alpha(x_i)^T \phi(x_j) \label{sim}
%\end{equation}
%where $\alpha(x)=W_{\alpha}x$ and $\phi(x)=W_{\phi}x$ are two linear transformations of the original ROI features $x\in \mathbb{R}^{d}$. The parameters $W_{\alpha}$ and $W_{\phi}$ are both $d\times d$ dimensional matrices and can be learned via back-propagation. Based on the Eq.~\ref{e} and Eq.~\ref{sim}, the similarity matrix $G$ is defined below.
%\begin{gather}
%\label{G}
%G_{i,j}=
%\begin{cases}
%Sim(x_i,x_j),& A_{i,j}=1\\
%-1e^{-10},& A_{i,j}=0.
%\end{cases}
%\end{gather}
%After obtain the similarity matrix with Eq.~\ref{G}, we perform normalization along each row of the matrix by using the following softmax function so that the sum of all the edges connected to the $i^{th}$ AU is 1. The final adjacency matrix $\widetilde{A}$ of the graph $G$ is:
%\begin{equation}
%    \widetilde{A}_{i,j} = \frac{\exp (G_{i,j})}{\sum_{j=1}^{C} \exp (G_{i,j})}. \label{adj}
%\end{equation}

%\subsubsection{Reasoning on Graph}

%Different from the traditional convolution which operates on a regular local grid region, graph convolution computes the response of each node based on its neighbor nodes defined by the adjacency matrix.
\begin{equation}
    Z=\sigma(\widetilde{A}XW),
\end{equation}
where $\widetilde{A}$ is the normalized adjacency matrix, $X$ is the stacked ROI features of AUs from different levels, and $W$ is the learnable weight matrix. $\sigma(\cdot)$ is an activation function. For each node in $G$, it first aggregates structure information from its neighbor nodes and then is mapped to the new state via $W$. %This layer-wise propagation can also be stacked into multi-layer GCNs to capture more complex structure relationships.
% and we choose ReLU in this paper

\subsection{MARGL Framework}
The backbone network, adaptive ROI learning module and multi-level AU relation embedding together compose the proposed MARGL framework. ResNet-18~\cite{he2016deep} is chosen as our backbone network. For different levels of the feature maps from ResNet, the initial sizes of cropped ROIs are set to be $10\times 10$, $5\times 5$ and $2\times 2$ as the global feature maps shrink. The cropped ROIs of all levels are then resized to $6\times 6$ before regional learning. Considering the symmetry of facial regions, two independent multi-level AU relation embedding blocks are employed respectively. We calculate the average of the enhanced ROI features from symmetrical face and employ fully connected layers to get the prediction. Predictions from three regional levels and global level are fused by element-wise maximum to obtain the final result. In order to better supervise the adaptive ROI feature learning across multiple levels, binary cross entropy loss function is applied to each regional branch. Then the final loss can be formulated as:
% due to the moderate parameter amount and computation cost while maintains good performance in object recognition tasks. The stride of the first convolution layer is modified to 1 to keep more fine-grained texture information of the facial image
%Note that for clarity we only draw one GCN in Fig.~\ref{framework}. The input and output feature dimensions of the graph inference are kept unchanged.

%\subsubsection{Loss Function}

%After we obtain the enhanced ROI features of different AUs, independent fully connected layers are applied to get prediction results. As shown in Fig.~\ref{framework}, we get three predictions from different levels of ROI features and one prediction from the global feature.

%Due to that AU recognition can be regarded as a multi-label binary classification problem, we adopt the binary cross-entropy loss for each branch. Meanwhile, the distribution of positive AU samples is extremely unbalanced in dataset. To ease the impact to the model, AU-wise weights are multiplied to the loss function and
\begin{equation}
    L =L_0(P_g, Y)+\sum_{k=1}^3 L_k(P_k, Y),
\end{equation}
where $L_k=-\frac{1}{C}\sum_{i=1}^C w_i[y_i \log p_i+(1-y_i) \log (1-p_i)]$. $P_k$ is the ROI-based prediction, $P_g$ is the global-based prediction, $Y$ is the ground truth. $w_i$ is introduced to ease the data imbalanced problem, and is determined by the occurrence rate of AU $i$ in the dataset. Specifically, $w_i=\frac{(1/r_i)C}{\sum_{i=1}^C (1/r_i)}$, where $r_i$ is the occurrence rate of AU $i$ in the dataset.

\section{Experiments}

\subsection{Experimental Setup}
We evaluate our method on two popular benchmark datasets, i.e., BP4D \cite{zhang2013high} and DISFA \cite{mavadati2013disfa}. For all experiments, we follow the protocol of subject independent 3-fold cross validation and report the F1-frame results for comparison which is widely used in the community.
%BP4D consists of facial expression data collected from 41 participants, including 23 females and 18 males. Each subject participated in 8 tasks which aims to induce different facial expressions. There are 146,847 frames with valid AU labels. DISFA is another spontaneous AU database which contains 27 videos recorded from 12 females and 15 males. There are totally 130,815 frames and each frame is labeled with AU intensity from 0 to 5. In DISFA experiments, frames with intensity equal or greater than 2 are treated as positive and the rest are negative.

%\subsection{Evaluation Metric}
%As AU recognition is a multi-label binary classification problem and samples in both datasets are very imbalanced, F1-frame, i.e., the harmonic mean of precision and recall, is commonly used for comparison. In this section, we report the F1-frame for each AU and the average F1-frame of all AUs.%, which is denoted as Avg in following tables.

%\subsection{Implementation Details}
For each frame, we first perform face detection and alignment based on similarity transformation and obtain a $200 \times 200$ RGB face image. After some common data augmentations, the facial images are resized to $192\times 192$ as the input of the network. During the testing phase, only center cropping is employed. Meanwhile, to locate the AU centers, facial landmarks are detected with open source toolbox Dlib~\cite{king2009dlib}. The ResNet-18 backbone takes ImageNet pre-trained model weight as initialization and the parameters of additional layers are initialized randomly. The model is trained via stochastic gradient descent (SGD) with the initial learning rate 0.01.
%a momentum of 0.9 in 10 epochs and the batch size is 64. The initial learning rate is 0.01 and decayed with the cosine learning rate schedule.All experiments are implemented on PyTorch~\cite{paszke2017automatic} platform.
%In order to enhance the diversity of training data, ordinary data augmentations such as randomly cropping and horizontally flipping are conducted.

\begin{table}[htbp]\small
	\centering
	\caption{Ablation study on BP4D dataset.}
	\label{ablation}
	\begin{adjustbox}{max width=\columnwidth} %\textwidth
		\begin{threeparttable}
            \begin{tabular}{c|ccccc|cccccccccccc|c}
            \hline
            Method & ROI & AROI & Multi-level & Intra-relation & Inter-relation & Avg\tabularnewline
            \hline
            \hline
            ResNet-18 &  &  &  &  &  & 60.8\tabularnewline
            +ROI & $\surd$ &  &  &  &  & 62.5\tabularnewline
            +AROI & & $\surd$ &  &  &  & 64.1\tabularnewline
            +MAROI & & $\surd$ & $\surd$ &  &  & 65.0\tabularnewline
            +AROI+Intra-level & & $\surd$ &  & $\surd$ &  & 64.8\tabularnewline
            +MAROI+Intra-level & & $\surd$ & $\surd$ & $\surd$ &  & 65.6\tabularnewline
            %Intuitive Multi-level & & $\surd$ & by pooling & $\surd$ &  & 64.7\tabularnewline
            MARGL & & $\surd$ & $\surd$ & $\surd$ & $\surd$ & \textbf{66.1}\tabularnewline
            \hline
            \end{tabular}
		\end{threeparttable}
	\end{adjustbox}
\end{table}

\subsection{Ablation Study}
We conduct ablation study on BP4D to verify the effectiveness of the proposed modules in our framework. Average F1-frame of 12 AUs under 3-fold cross validation are reported.

%\subsubsection{Impact of adaptive ROI learning}
To evaluate the effectiveness of adaptive ROI learning module, we compare it with the original ROI learning method where the ROI is cropped with fixed size and position rules. As presented in Table~\ref{ablation}, the baseline model, i.e., ResNet-18, achieves the average F1-frame of 60.8\%. After adding the original ROI module (+ROI), the average performance is improved by 1.7\%. Nevertheless, with the proposed adaptive ROI module (+AROI), the location and size of bounding box of ROI can adjust adaptively and further leads to 1.6\% improvement compared to original version, which demonstrates more discriminative ROI features are captured by AROI. The visualization of original and adaptive ROIs of AU4 and AU12 are illustrated in Fig.~\ref{fig1} (a) and (b). After adaptive ROI learning, the bounding boxes of AU4 ROIs are narrowed down to focus on the inner part of brows, while the ROIs of AU12 are expanded to the wrinkle regions caused by the lip movement.

%After cropping, individual convolutional layers are applied to further refine the regional features. As the module can be inserted at any level of the backbone network, we conduct experiments to insert it after conv3, conv4 and conv5 respectively. Best experimental results are reported and the best insertion place is after conv3, which is probably due to that conv3 feature maps can offer more regional texture details since the global feature maps already provide semantical information.

%The module is inserted after the third block of convolutional layers as these feature maps can offer more regional texture details.

%\subsubsection{Impact of multi-level AU relation embedding}
%To demonstrate the effectiveness of multi-level AU relation embedding, we conduct ablation study by decomposing it into the multi-level manner and relation embedding.

Furthermore, we insert the AROI learning module to the last three levels of backbone network and investigate the effectiveness of multi-level manner (+MAROI). From Table~\ref{ablation}, nearly 1\% improvement is gained due to richer regional features and more supervision information. For relations between AUs, we conduct experiments in both single-level and multi-level manner. With respect to single-level manner (+AROI+Intra-level), only the intra-level graph of a single level is employed and the overall performance is improved by 0.7\%. Similarly, incorporating the prior relation knowledge in multi-level manner (+MAROI+Intra-level) leads to 0.6\% improvement. These results indicate that AU relationship knowledge can contribute to the recognition performance. Moreover, taking the inter-level AU regional relevance into consideration, we obtain the complete MARGL model and achieve the average F1-frame of 66.1\%. Compared to the one without encoding the inter-level relation, the result is improved by 0.5\%, which demonstrates the correlation information between ROI features across levels can be leveraged to boost the model performance. %It is notable that with multi-level adaptive ROI and graph learning, we boost the performance by 5.3\%.

%Meanwhile, we also experiment with another intuitive way to build multi-level structure (Intuitive Multi-level). Referring to~\cite{li2019}, feature maps of previous levels are pooled to the same size and stacked together. Then adaptive ROI learning and intra-level AU relation knowledge embedding are employed. The performance is 0.9\% worse compared to our multi-level manner which verifies our method is more effective to utilize the multi-level information.

\begin{table}[htbp]\small
	\centering
	\caption{Comparison of F1-frame on BP4D dataset.}
	\label{bp4d}
	\begin{adjustbox}{max width=0.78\columnwidth}
		\begin{threeparttable}
            \begin{tabular}{c|ccccccc}
            \hline
            \multirow{1}{*}{AU} & DRML & ROI & JAA & DSIN & SRERL & MARGL\tabularnewline
            \hline
            \hline
            1 & 36.4 & 36.2 & 47.2 & 51.7 & 46.9 & \textbf{59.0}\tabularnewline
            2 & 41.8 & 31.6 & 44.0 & 40.4 & 45.3 & \textbf{53.4}\tabularnewline
            4 & 43.0 & 43.4 & 54.9 & 56.0 & 55.6 & \textbf{58.8}\tabularnewline
            6 & 55.0 & 77.1 & 77.5 & 76.1 & 77.1 & \textbf{79.5}\tabularnewline
            7 & 67.0 & 73.7 & 74.6 & 73.5 & 78.4 & \textbf{79.8}\tabularnewline
            10 & 66.3 & 85.0 & 84.0 & 79.9 & 83.5 & \textbf{85.1}\tabularnewline
            12 & 65.8 & 87.0 & 86.9 & 85.4 & 87.6 & \textbf{88.8}\tabularnewline
            14 & 54.1 & 62.6 & 61.9 & 62.7 & 63.9 & \textbf{66.1}\tabularnewline
            15 & 33.2 & 45.7 & 43.6 & 37.3 & 52.2 & \textbf{57.7}\tabularnewline
            17 & 48.0 & 58.0 & 60.3 & 62.9 & \textbf{63.9} & 62.2\tabularnewline
            23 & 31.7 & 38.3 & 42.7 & 38.8 & 47.1 & \textbf{48.8}\tabularnewline
            24 & 30.0 & 37.4 & 41.9 & 41.6 & 53.3 & \textbf{54.2}\tabularnewline
            \hline
            Avg & 48.3 & 56.4 & 60.0 & 58.9 & 62.9 & \textbf{66.1}\tabularnewline
            \hline
            \end{tabular}
		\end{threeparttable}
	\end{adjustbox}
\end{table}

\begin{table}[htbp]\small
	\centering
	\caption{Comparison of F1-frame on DISFA dataset.}
	\label{disfa}
	\begin{adjustbox}{max width=0.78\columnwidth}
		\begin{threeparttable}
            \begin{tabular}{c|ccccccccc}
            \hline
            AU & DRML & ROI & JAA & DSIN & SRERL & MARGL\tabularnewline
            \hline
            \hline
            1 & 17.3 & 41.5 & 43.7 & 42.4 & 45.7 & \textbf{54.4}\tabularnewline
            2 & 17.7 & 26.4 & 46.2 & 39.0 & \textbf{47.8} & 45.7\tabularnewline
            4 & 37.4 & 66.4 & 56.0 & 68.4 & 59.6 & \textbf{70.8}\tabularnewline
            6 & 29.0 & \textbf{50.7} & 41.4 & 28.6 & 47.1 & 42.8\tabularnewline
            9 & 10.7 & \textbf{80.5} & 44.7 & 46.8 & 45.6 & 60.1\tabularnewline
            12 & 37.7 & \textbf{89.3} & 69.6 & 70.8 & 73.5 & 75.8\tabularnewline
            25 & 38.5 & 88.9 & 88.3 & 90.4 & 84.3 & \textbf{94.7}\tabularnewline
            26 & 20.1 & 15.6 & 58.4 & 42.2 & 43.6 & \textbf{65.8}\tabularnewline
            \hline
            Avg & 26.7 & 48.5 & 56.0 & 53.6 & 55.9 & \textbf{63.8}\tabularnewline
            \hline
            \end{tabular}
		\end{threeparttable}
	\end{adjustbox}
\end{table}

\subsection{Comparison with State-of-the-Art Methods}
We compare the proposed MARGL with previous state-of-the-art methods under the same protocol. The competitive approaches include DRML~\cite{zhao2016deep}, ROI~\cite{li2017action}, JAA-Net~\cite{shao2018deep}, DSIN~\cite{corneanu2018deep} and SRERL~\cite{li2019}.

Table~\ref{bp4d} shows the experimental results on BP4D dataset. Except for AU17, the proposed MARGL outperforms all other competing methods. Compared to region learning approaches, such as ROI and JAA-Net, which make predictions with pre-defined AU regions, the adaptive ROI learning in our method further boosts the performance. Compared with SRERL which involves AU relationship inference with GGNN, our model achieves 3.2\% improvement in average F1-frame which indicates the effectiveness of the intra and inter level relations. Experimental results on the more challenging DISFA dataset are shown in Table~\ref{disfa}. Compared to the previous state-of-the-art methods, MARGL achieves a significant improvement of 7.8\% and obtains 63.8\% in terms of average F1-frame. Due to the modules proposed in the framework, MARGL works well for majority AUs and maintains consistently satisfying performance.
%As DISFA dataset contains less AU samples and the data imbalance problem is very severe, a relatively big drop in performance exists in most of the previous methods. However,

\section{Conclusion}
In this paper, we propose a novel multi-level adaptive ROI and graph learning method for AU recognition. An adaptive ROI learning module is proposed to automatically adjust the location and size of ROI for each AU in a data-driven manner. Meanwhile, multi-level AU relation graph is constructed to model the intra-level AU relation and inter-level AU regional relevance jointly to further enhance the discrimination of ROI features. State-of-the-art performances achieved on BP4D and DISFA demonstrate the superiority of our method.

%\section{Acknowledgment}
%This work is supported by National Key Research and Development Project of China (2018YFC0807702).

\vfill\pagebreak
%\newpage

\bibliographystyle{IEEEbib}
\bibliography{refs}

\end{document}